\documentclass[10pt,amssymb,twocolumn,notitlepage]{article}

\usepackage{epigraph}
\usepackage[affil-it]{authblk}
\usepackage{dcolumn}
\usepackage[stable]{footmisc}
\usepackage{blkarray}
\usepackage{bm}
\usepackage[mathscr]{euscript}
\usepackage[font=scriptsize]{caption}
\usepackage{subcaption,graphicx}
\usepackage{amsmath}
\usepackage[table]{xcolor}
\usepackage[margin=1.7cm]{geometry}
\usepackage{tcolorbox}
\usepackage{hyperref}
\usepackage{gensymb}
\usepackage[font=footnotesize]{caption}
\usepackage{hyperref}
\usepackage{amssymb}
\usepackage{makecell}

\title{Toward Generalized Clustering through an \\ One-Dimensional Approach}
\author{Luciano da Fontoura Costa \\ \emph{luciano@ifsc.usp.br}}
\affil{S\~ao Carlos Institute of Physics -- DFCM/USP, Brazil} 

\begin{document}

\twocolumn[
\begin{@twocolumnfalse}
    \maketitle
    \begin{abstract}
    After generalizing the concept of clusters to incorporate clusters that are linked to other
    clusters through some relatively narrow bridges,
    an approach for detecting patches of separation between these clusters  
    is developed based on an agglomerative clustering, more specifically
    the single-linkage, applied to one-dimensional slices obtained from respective feature spaces.   
    The potential of this method is illustrated with respect to the analyses of
    clusterless uniform and normal distributions
    of points, as well as a one-dimensional clustering model characterized by
    two intervals with high density of points separated by a less dense interstice.  This
    partial clustering method is then considered as a means of feature selection and cluster identification,
    and two simple but potentially effective respective methods are described and illustrated with 
    respect to some hypothetical situations.
    \end{abstract}
\end{@twocolumnfalse} \bigskip
]

\setlength{\epigraphwidth}{.49\textwidth}
\epigraph{`Ogni blocco di pietra ha una statua dentro di s\'e ed \`e compito dello scultore scoprirla.'}{\emph{Michelangelo.}}

\section{Introduction}

Grouping entities into abstract categories, or \emph{clustering} (e.g.~\cite{Koutrombas,DudaHart,Kotsiantis}), constitutes one of the most  fundamental and intrinsic human activities.  By doing so, it is possible to 
avoid the explosion of labels that would be otherwise implied by individual identification of each possible 
entity.  The often considered basic grouping rationale is that the entities in each category would share several 
features, while differing from entities in other categories.  So, emphasis is placed on identifying these 
common and distinguishing features.  For simplicity's sake, we will call this basic type of groups
as \emph{granular clusters}.

Continuing application of clustering principles by humans ultimately gave rise to language, arts, and
scientific modeling.  Indeed, each word can be understood a model corresponding to a respective
category of objects, actions, etc.~(e.g.\cite{CostaModeling}).  
The great importance of clustering, as well as the challenging
of achieving it computationally, has been reflected in the relatively large number of related 
approaches reported in the literature (e.g.~\cite{Koutrombas,DudaHart,Kotsiantis}).

Many of these approaches have been based, or relate to, the above mentioned basic
grouping principle.  In other words, it is often expected that, when mapped into feature spaces,
distinct categories will give rise to relatively well-separated compact groups of points.
In other words, large inter-group scattering and small intra-group scattering are generally
expected.  

Another related, but somewhat complementary approach is to understand as
clusters groups of points that can be well-separated through hypersurfaces, or separatrices.
These two approaches differ in the important sense that the existence of a separatrix does
not necessarily require the granular principle, while the latter usually ensures the former.
In a sense, the separatrix principle can then be understood as being less strict than the granular
counterpart.

In both cases, it is often assumed that each cluster is \emph{completely segregated} from all the
other clusters, therefore constituting \emph{separated}, \emph{isolated} groups.  Even when
some level of overlap is present, frequently one still aims at obtaining completely separated
clusters. This not so often realized assumption implies an important characteristic, namely 
that the cluster identification is a
\emph{global} activity, in the sense that the identification of groups or separations between them
is performed while taking into account the whole `border' of each group, along all possible
directions.  In other words,
this type of requirement acquires a topological aspect regarding the isolation between the involved
clusters.  Though contributing to a more complete characterization of the clusters, this requirement
implies conceptual and computational demands that are often hard to be met by respective 
algorithms.  

One of the motivations of the present work is to allow groups of points that are evidently not
separated from the remainder groups to be considered as a kind of generalized clusters,
characterizing by the existence of \emph{incomplete separation between} the respective
groups, in the sense that two groups can be linked even through `solid', but relatively narrow
bridges, while still being separated along longer border extensions. These clusters are henceforth called,
paradoxically, \emph{linked clusters}, corresponding to a generalization of the concept
of cluster.  As a consequence, focus needs to be placed on \emph{identifying the existing 
separation extensions}, even if in terms of respective \emph{patches}.

The basic rationale is that \emph{any} identification of separation patches between
two adjacent groups is intrinsically important, even if incomplete.   Indeed, the complete
segregation of clusters can be though as the integration of many separation patches.
A related approach, known as multi-view clustering (e.g.~\cite{BickelScheffer}), considers 
separating sets of features (multi-views) as the means of improving clustering.
Here, we resort to slicing the feature space through narrow hypercylinders aligned along each of the existing 
features as the means of obtaining separation patches between clusters, which can provide
valuable information about the relationships between sets of points in the original feature space.

Patched clustering approaches can also be used to devise potentially 
effective methods for feature selection,  in the sense that if a given feature is  found to contribute to
several local separations, it would also be a good candidate to contribute to respective
complete segregations.   

The pathed clustering approach allows some interesting features, such as the possibility to
search for partial separations by considering lower dimensional samples of the original
feature spaces. Though other related approaches could be adopted, here we focus on
the idea of one-dimensional slicing (or probing) of feature spaces through hypercylinders as a means for identification
of patches of separation between clusters.
This approach is potentially interesting because of its relative low computational complexity, as
well as for its ability to probe the feature space clustering structure in a more `surgical' and
independent way, tending to avoid interactions and projections between clusters that could otherwise 
be more intense in the case of higher-dimensional approaches.  

For all these reasons, the present work develops the concepts of partial separation as a way to
performing generalized clustering.  Observe that the concepts of partial clusters and one-dimensional
identification separating interstices can be thought as going hand in hand.

This article starts by discussing the concept of generalized clustering to include linked clusters, 
and follows by presenting the adopted one-dimensional patched clustering
approach, which involves the single-linkage agglomerative method.  Then, the application
of these concepts to deriving a simple and yet potentially effective feature selection methodology
and  to develop of a simple local clustering algorithm are then presented and illustrated.

\section{Generalized Clusters and Separations}

\begin{figure}[h!]  
\centering{
\includegraphics[width=4cm]{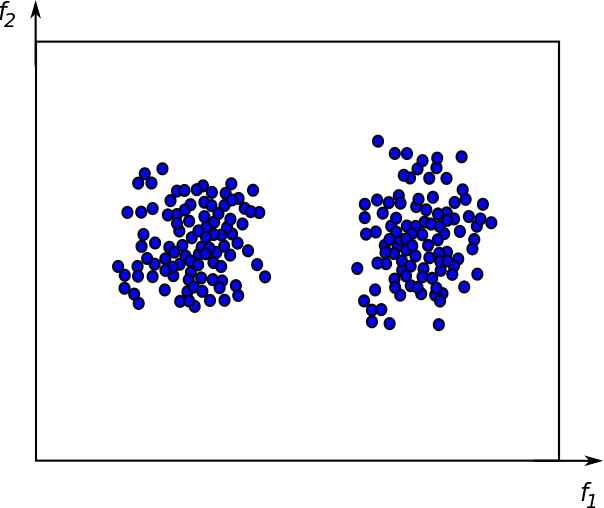}  
\includegraphics[width=4cm]{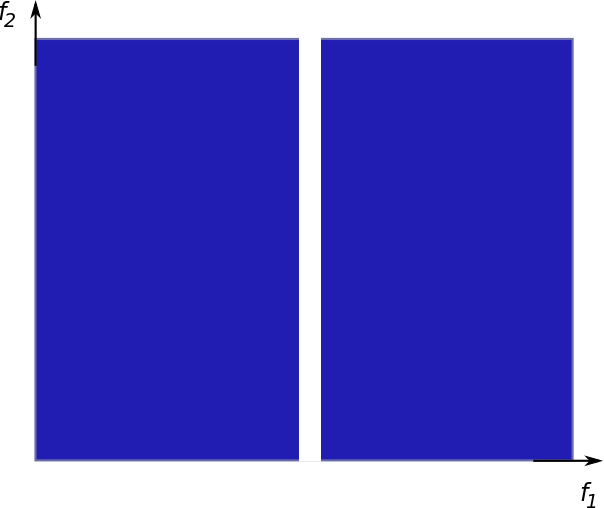}  \\
(a)  \hspace{3.5cm} (b) \\
\includegraphics[width=4cm]{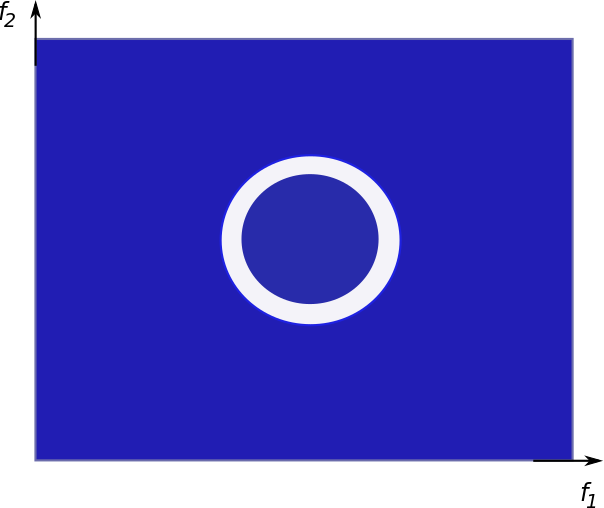}  
\includegraphics[width=4cm]{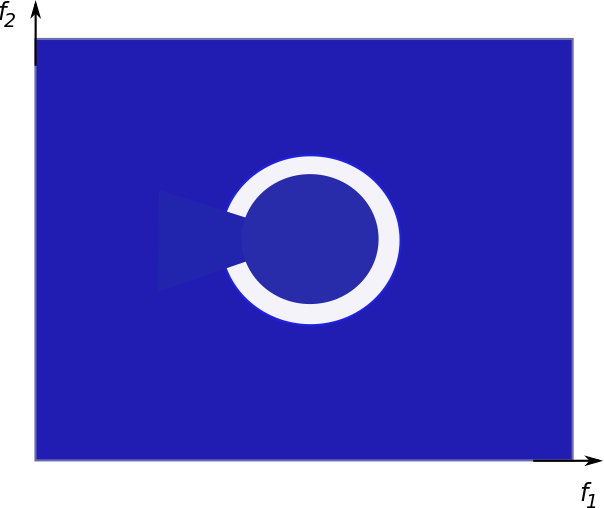}  \\
(c)  \hspace{3.5cm} (d) \\
\includegraphics[width=4cm]{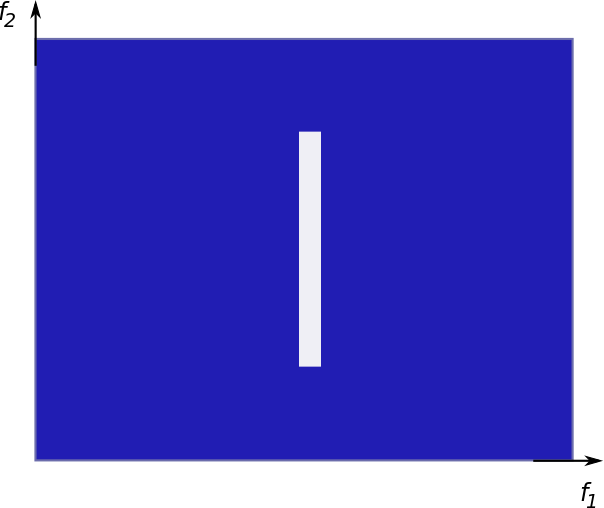}  
\includegraphics[width=4cm]{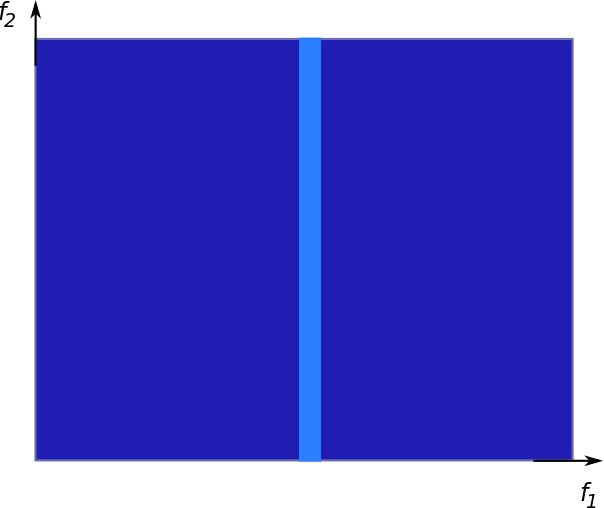}  \\ 
(e)  \hspace{3.5cm} (f) 
\caption{Six distinct types of clusters and respective separations:
(a) granular cluster; (b) separatrix cluster; (c) surrounded cluster
(a special case of separatrix cluster); (d) a linked cluster; (e)
another type of linked cluster; and (f) a separatrix cluster with
fuzzy separation.  The continuous areas can also be understood
as corresponding to respective distributions of points.}
\label{fig:exs}}
\end{figure}

The first central issue regards what we understand by a \emph{cluster}  or \emph{separation},
which tend to be dual concepts. 
We have already observe that, in the respective literature, a cluster is often defined as a set 
of entities that are similar one another while being different from other entities, as illustrated
in Figure~\ref{fig:exs}(a), which we have called a \emph{granular cluster}.  All the examples
in this figure assume a two-dimensional feature space defined by hypothetical
measurements $f_1$ and $f_2$.

Another type of cluster already mentioned is related to the concept of separatrix.  For instance,
the two clusters in Figure~\ref{fig:exs}(b) are not granular, but can be divided by a 
separatrix.  We have called this type of cluster a \emph{separatrix cluster}.  Recall that
granular clusters are almost invariably separatrix clusters, but not vice-versa.   Also,
observe that the two cluster examples discussed so far can be properly separated by
projections along some direction (in the case into the horizontal axes).

Figure~\ref{fig:exs}(c) illustrates another type of separatrix cluster that is not a granular
cluster and that cannot be isolated by a linear separatrix: the inner cluster is completely
surrounded by the other cluster.   As a consequence, it becomes impossible to try 
identifying the separation through consideration of projections along any direction.

A particularly interesting situation is depicted in Figure~\ref{fig:exs}(d).  Here, we have a 
situation similar to the previous example, but the inner cluster turns out to be linked
somehow to the outer cluster through a narrow bridge.  This type of cluster is henceforth
called a \emph{linked cluster}.  Though, properly speaking, this could not be taken to be a
cluster, there are several reasons for extending that concept for such situations.  One of
them is that the bridge could correspond to an artifact implied by
noisy or unsuitable features.  Another reason is that the cluster is indeed intrinsically
linked, but that this link has transient nature or is likely to be removed, or even, that
has just been created.  Yet another reason is that, even if indeed permanently attached,
the inner group of points can still be considered mostly distinct and different from the outer
cluster.  

Figure~\ref{fig:exs}(e)  illustrates another example of linked cluster.  Though the left and
right-hand sides of the points are topologically linked, it is still interesting to know
that the points to each side of the slit are \emph{locallly} disconnected, which can provide
valuable insights about the situation under analysis.

We can also have that the separation interstices are not completely void, as in the cases
shown in Figure~\ref{fig:exs}(a-e), but exhibit an intermediate density of points, giving
rise to fuzzy separations that can be related to overlaps between clusters.

In addition to the above types of clusters, we could also have hybrids involving 
combinations of these types.   For instance, we can have a circular inner cluster 
whose border involves void separation, fuzzy patches, as well as bridges.  
The possibility of coexisting different types of clusters and separations corroborates 
the difficulty of devising global clustering approached
capable of identifying this type of cluster.

In the present work, we aim at considering and identifying all these types of clusters and
respective separations, yielding a more generalized approach to clustering. 
In addition to providing valuable information about the relationships between the points in 
different regions of the feature space, patched identification of clusters/separations can also
be understood as a preliminary step to a more global approach, where the locally identified
clusters/separations are integrated into larger, potentially complete clusters and separatrices.

\section{One-Dimensional Clustering}

The issue of one-dimensional clustering, though intrinsically fundamental and interesting,
has received relatively little attention when compared to multi-dimensional counterparts.
One possible reason is that the characterization of entities can rarely be comprehensive
when just one measurement is adopted.   However, it follows from the previous discussions
that one-dimensional exploration of entities mapped into feature spaces is intrinsically
valuable as a subsidy for finding relevant features and patched indications of clustering and
interstices.  More specifically, every indication of clustering, even if detected by a 1D slicing of
the feature space, is important and welcomed.

In this section, we discuss the problem of one-dimensional clustering in terms of agglomerative
approaches (e.g.~\cite{DudaHart}), in particular the single-linkage method.  Compared to other methods 
such as $k-$means, agglomerative methods have an intrinsic desirable feature, namely its inherent 
ability to represent and characterization data separation in terms of several spatial scales,
yielding a respective dendrogram from which the data can be divided in any number of categories (by
considering specific clustering heights).
This allows not only a more complete characterization of the clustering along these scales,
but also provides the means for identifying the relevance of each potential cluster in terms
of the length of its respective branch.  In 
addition, adaptive schemes can be considered in which the resolution of the separation
can be progressively increased along spatial scales, such as zooming (e.g.~\cite{Mucha}).

In the single-linkage method, groups are progressively merged in terms
of the nearest distance between them. Recall that the minimum distance between two sets of
points corresponds to the minimum distance between any of respective pairs of points.
The choice of this method, which is know to be susceptible to the phenomenon of
\emph{chaining}, among many other agglomerative possibilites, was justified by
preliminary experiments in which the single linkage tended to be more likely to reject
clusters in the case of uniform and normal distributions of points.  In addition, this method
is particularly simple regarding conceptual and computational aspects.

A discussion of one-dimensional clustering can benefit greatly from having models of cluster
structures in one-dimension.   We start by considering a situation not involving clustering
(other than for statistical fluctuations), namely the case of points scattered in a randomly uniform
manner along a one-dimensional domain $x$.  

It can be shown that the nearest neighbor distance between such points is given by the
exponential probability distribution 

\begin{equation} \label{eq:pd}
 p(d) = \lambda e^{-\lambda d}
\end{equation}

whose average is $1/\lambda$.

Let's consider the single linkage method, which progressively joins clusters based on the
their nearest distance, corresponding to the smallest distance between any point of each
pair of clusters.   Interestingly, because of the adjacency implied by 1D domains, we have
that merging can only take place among adjacent clusters.  As a consequence, the successive 
differences of clustering height also follows Equation~\ref{eq:pd}.

\begin{figure}[h!]  
\centering{
\includegraphics[width=8cm]{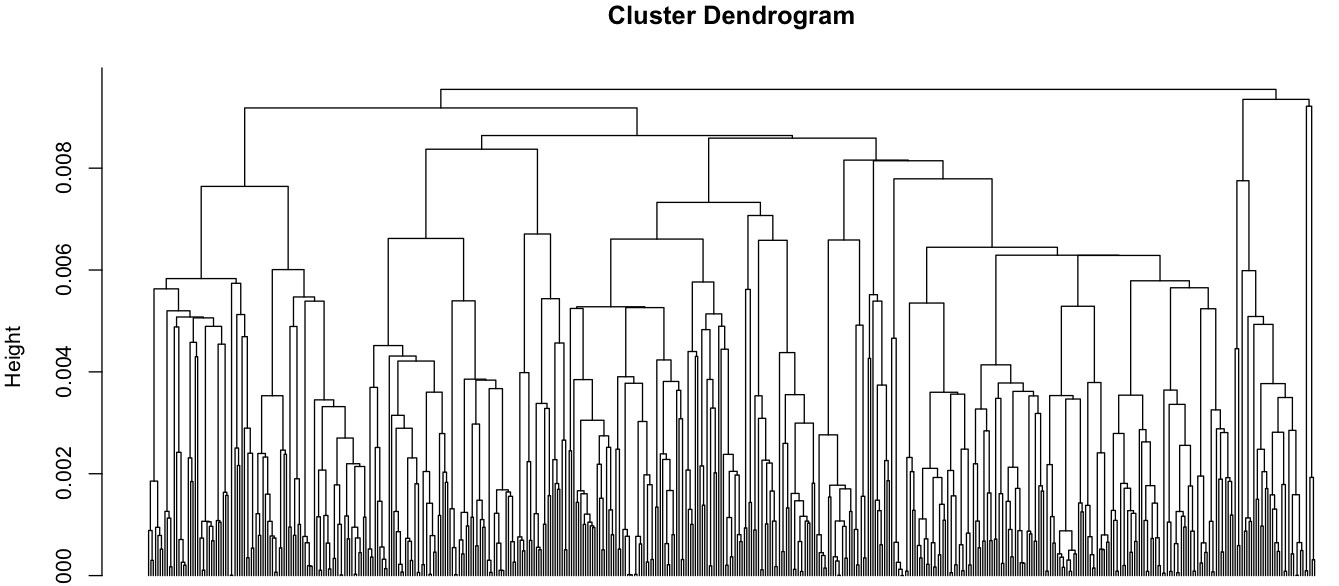}  
\caption{Dendrogram obtained by single-linkage agglomerative clustering of a set of points
with $\lambda = 500$ uniformly distributed along the one-dimensional internval $0 \leq x \leq 1$
Observe the absence of any relevant cluster indications.}
\label{fig:unif}}
\end{figure}

Figure~\ref{fig:unif} shows the dendrogram obtained by single-linkage agglomerative
clustering of a uniformly scattered set of points with $\lambda = 500$ distributed in 
the interval $0 \leq x \leq 1$.

As expected, no significant cluster is suggested by this dendrogram, given the relatively
short extension of the the branches leading to relatively large groups of points.  Also, observe
that the total agglomeration is reached at about 5 times the average expected nearest
neighbor distance of $1/\lambda = 1/500 = 0.002$.

\begin{figure}[h!]  
\centering{
\includegraphics[width=8cm]{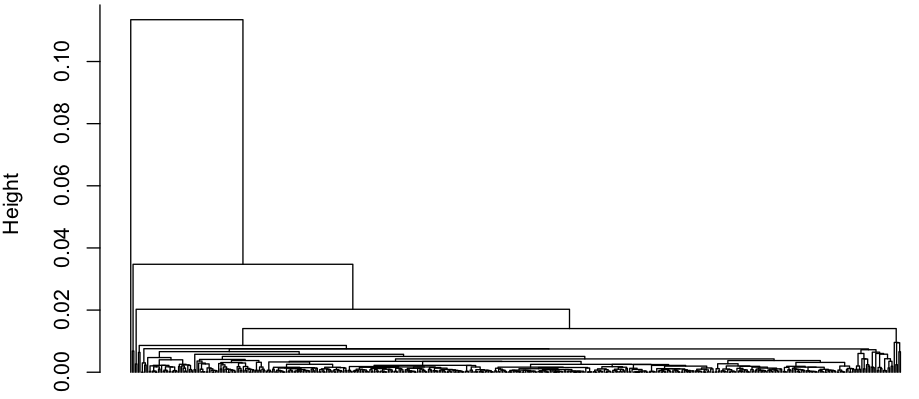}  
\caption{Dendrogram obtained by single-linkage agglomerative clustering of a set of points
with $\lambda = 500$ normally distributed with zero means and standard deviation equal to $0.1$.
The dendrogram, which exhibits the chaining effect, provides no indication about any relevant cluster.}
\label{fig:normal}}
\end{figure}

Figure~\ref{fig:normal} depicts the dendrogram obtained for a normally distributed set of points
with means equal to zero and standard deviation equal to $0.1$.  This type of points distribution
is often observed in practice, being characterized by a gradual increase of point density near
the mean.  The obtained dendrogram exhibits the chaining effect and does not
provide any indication of relevant clusters.

\begin{figure}[h!]  
\centering{
\includegraphics[width=8cm]{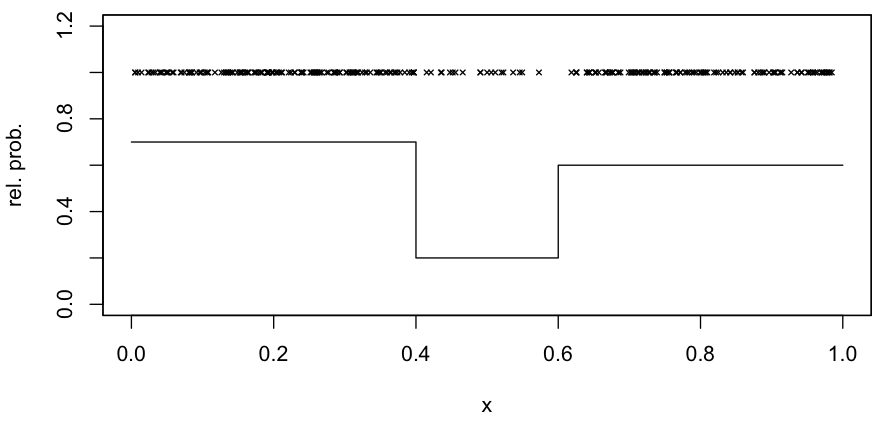}  
\caption{Scattering of points (above), containing two clusters, obtained by sampling the three intervals in the 
adopted model with $\lambda_1 = 140$, and $\lambda_2 = 20$ and $\lambda_3 = 120$, which
are associated to the uniform probability density shown below.}
\label{fig:ex_model}}
\end{figure}

Let's now consider a \emph{prototype} of one-dimensional cluster consisting of two regions
of points uniformly distributed with larger intensities $\lambda_1$ and $\lambda_3$ along 
respective intervals $0 \leq x < x_1$ and  $x_2 \leq x < 1$, while a more sparse distribution with 
$\lambda_2 < \lambda_1$ is implemented in the interval $x_1 \leq x < x_2$.   Figure~\ref{fig:ex_model}
illustrates this configuration.

\begin{figure}[h!]  
\centering{
\includegraphics[width=8cm]{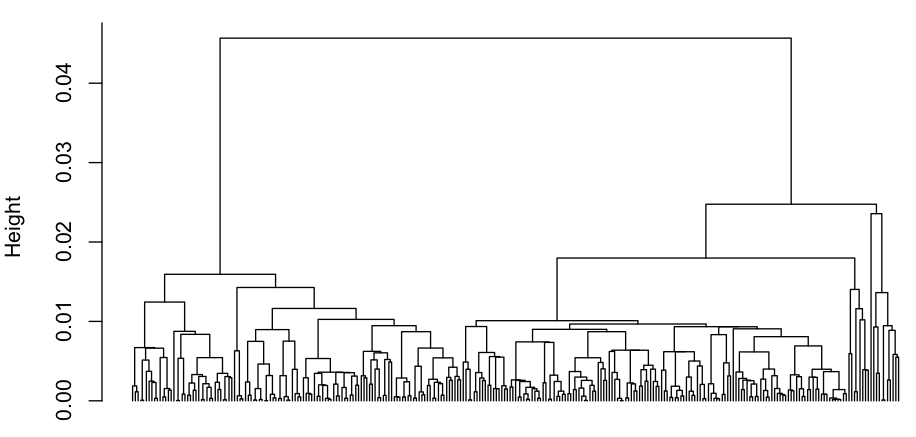}  
\caption{The dendrogram obtained by single-linkage clustering of a the considered one-dimensional
cluster structure shown in Figure~\ref{fig:ex_model}.  The two main clusters are effectively identified despite the
relatively small separation $\Delta x= x_2 - x_1 = 0.2$. Observe the streaks at the right-hand side, 
which corresponds mostly to the points in the interstice.  The relevance of existence of clustering
in this dendrogram obtained by using the proposed algorithm was $\rho\approx 0.7453$, with the length
of the second longest branch, streaks ignored, corresponding to $\approx 0.034$.}
\label{fig:dendr_model}}
\end{figure}

The two groups of points arises as a consequence of the less intense distribution of
points existing in the intermediate interval with length $\Delta x = x_2 - x_1$, which we will henceforth
call an \emph{interstice}, among the two groups corresponding to the other two denser intervals.
Figure~\ref{fig:ex_model} illustrates one example of this clustering model with respect to
$x_1 = 0.4$, $x_2 = 0.6$, $\lambda_1 = 140$, and $\lambda_2 = 20$ and $\lambda_3 = 120$.
The respectively obtained dendrogram, by using single-linkage, is shown in Figure~\ref{fig:dendr_model}.

Unlike the dendrogram obtained previously, now we have an evident indication of the existence of
two clusters, characterized by long branches leading to groups with a substantial number of points, corresponding
to the two expected clusters.  Observe that the lenghth of the two main clusters is much larger than the
average nearest neighbor distance of $1/280 \approx 0.003571$ that would have been observed for a 
completely uniform scattering of $280$ points in the interval $0 \leq x < 1$.

Though we develop our two-modal clustering model based on uniform distributions
of points, similar properties could be expected when each cluster follows
other types of statistical distributions, such as normal.
Also, observe that this models considers the situation in which the interstice is not completely void.  

The points belonging to interstice tend to appear as long, narrow branches leading to
relatively small number of points.  These branches, which we will henceforth call \emph{streaks}, 
need to be ignored while detecting the more relevant clusters.   There are several possible means
to do so, but here we resort to the following simple method:

Starting from the root ($k=2$ clusters), cut the dendrogram at successive number of groupe $k$.  For each
of these cases,  consider
only those branches leading to at least $n = N/\alpha$ points, where $N$ is the total number of points
and $0 < \alpha \leq 1$ is a parameter.  This allows most of the streaks to be pruned.
Stop when the number of remaining branches is smaller than 2.  The length $\ell$ of the cluster
at the interruption time is  divided by the largest height value $H$ and taken as an indication of the \emph{relevance} $\rho = \frac{\ell}{H}$ of the existence of clustering in the
original set of points, with $0 \leq \rho \leq 1$.

Observe that the above algorithm tends to work even if more than two main well-defined 
clusters are present  in the original data, provided they have size larger than $n$.

By using this algorithm with $\alpha = 4$, we obtain a relevance $\rho \approx 0.7453$ for the 
dendrogram in  Figure~\ref{fig:dendr_model}.

The potential of this simple approach has been supported in all the examples above, and also by many
other configurations not included in this article.  Also, observe that this method has potential for
properly identifying all the types of clusters/separations in Figure~\ref{fig:exs}.  This can be in part
understood as a consequence of the ability of the adopted one-dimensional clustering approach
to emphasize the existence of two adjacent
concentrated densities of points, being relatively less affected by the type of separation.  Another
reason for the potential effectiveness of this method is its ability to focus on local separations,
therefore avoiding interferences of points in other regions or orientations, something that can
hardly be achieved by methods based on projections.

The above discussion and results suggests that this method can be adopted for
two important tasks, namely feature selection and cluster detection, which are respectively addressed 
in the following sections.

\section{Feature Selection}

Feature selection ( e.g.~\cite{LiuMotoda,ZhengCasari}) is important both for 
supervised and unsupervised pattern recognition.
Even in deep learning, when feature selection is often understood to be part of the learning dynamics
and not expected to be pre-defined, the identification of particularly effective features can be of 
interest while trying to understand how the classification was achieved.

Basically, given $M$ features of any type, feature selection aims at identifying those that contribute
more decisively to the proper classification of the existing groups.  As a consequence of the importance
of feature selection, a relatively large number of approaches has been reported in the literature
(e.g.~\cite{LiuMotoda,ZhengCasari}).

Two main types of feature selection approaches are often identified: (i) those which consider the
result of the classification itself as an indication of the relevance of specific sets of features, which
is often called \emph{wrapper}; and (ii)
those, called \emph{filter}, in which this relevance is inferred by indirect methods, such as in terms
of scattering distances between the existing clusters or correlations between the features.

Here, we develop a simple filter approach in which each of the $M$ features defines respective 
one-dimensional slicings of the feature space.  For generality's sake, and in order to consider more 
information and more points in each slice than could be otherwise obtained infinitesimally, 
we consider that each of these straight slices correspond to hypercylinders with radius $r$
in the respective $M$-dimensional feature space.

More specifically, the simple feature selection approach suggested here consists of:  for each feature 
$f_i$, $i = 1, 2, \ldots, M$, fix all other feature values as $f_j= \tilde{f}_j = constant$, $j \neq i$,
which defines the line (the hypercilinder axis)
$L_i: (f_1= \tilde{f}_1, f_2 = \tilde{f}_2, \ldots , f_i, \ldots, f_M =
\tilde{f}_M)$, with $f_{i,min} \leq f_i \leq f_{i,max}$.   Identify all marked points in the feature space 
that are at a maximum distance $r$ from $L_i$, given as

\begin{equation}
  r = \sqrt{ \sum_{\substack{j=1, \\ j \neq i}}^{M}  \left( f_j-\tilde{f}_j \right)^2 }
\end{equation}

The $f_i-$coordinates of each of these points define the one-dimensional signal $x$ to be
analyzed.

Relatively high relevance values $\rho$ obtained at least for some instances of the one-dimensional
sliced signals $x$ , can be potentially taken as indication of feature $f_i$ being relevant for
consideration in respective clustering approaches.

As a consequence of its `surgical' operation in the feature space, this method is potentially 
capable of estimating the contribution that each feature can provide even with respect to 
less frequently considered
types of interstices/borders between clusters.  In addition, given that it does not involve global projections,
the potential of each feature can be better assessed regarding its possible contribution to separating
groups.

In order to illustrate the potential of this simple approach, we consider the situation in which
we have squares with side $\gamma$ and circles with radius $\gamma$, with $\gamma$ 
uniformly distributed in the interval $1 \leq \gamma \leq 2$.   

Let's consider the following 5 features for the characterization of these shapes:
(i) $\gamma$, uniformly distributed for both circles and squares; (ii) perimeter ($p=2 \pi s$ for circles and
$p=4 s$ for squares; (iii) area ($a = \pi s^2$ for circles and $s^2$ for squares); (iv) relative perimeter
$r = p/s$ ($2 \pi$ for circles and $4$ for squares); and (v) circularity $c = \frac{4 \pi a}{p^2}$ ($1$ for
circles and $\frac{\pi}{4}$ for squares).  The two latter measurements are expected to contribute more
effectively to the separation between circles and squares, as they do not depend on their respective
scales defined by respective varying values of $\sigma$ and present different values for the two types 
of shapes.

Each of these obtained features was respectively standardized, i.e.

\begin{equation}
  \text{new feature} = \frac{\text{feature - (mean of feature)}}{\text{(standard deviation of feature)}}
\end{equation}

which ensures that each normalized feature has means equal to zero and standard deviation
equal to 1.  In addition, most of the normalized values fall within the interval $[-2,2]$, which 
allows us to define the region of interest in the feature space as $-2 \leq f_i \leq 2$ for
$i=1, 2, \ldots, M$.

Uniformly distributed noise in the range $[-0.2,0.2]$ was added to each feature, after standardization.
The slices were taken for all permutations of uniformly spaced feature values of $-2,-2,0,1,2$
and the radius of the hypercylinder was set as $r=2$, and $\alpha = 4$.  Only slices $x$ containing
more than $150$ points were considered by the one-dimensional clustering approach.

\begin{figure}[h!]  
\centering{
\includegraphics[width=9cm]{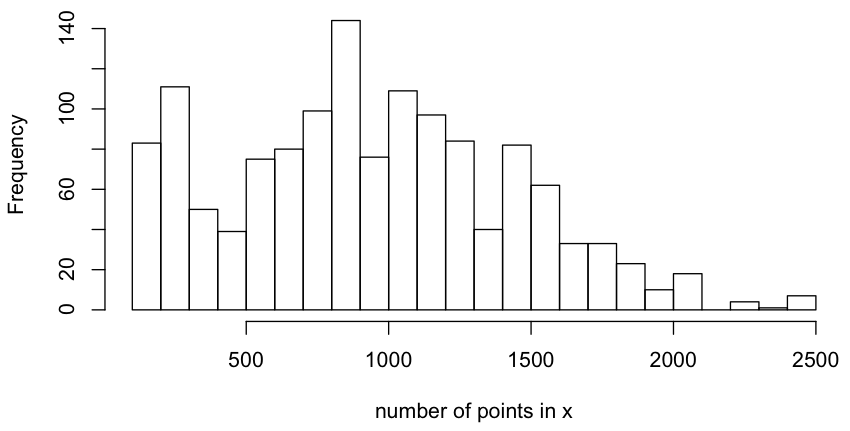}  
\caption{Histogram of the number of points in the analyzed one-dimensional slices of the 
feature space containing circles and squares.}
\label{fig:hist_sizes}}
\end{figure}

Figure~\ref{fig:hist_sizes} shows the histogram of the number of points in each of the sliced
$x$ signals, with an average of $940.34$.  We have that the number of points in 
most of the analyzed one-dimensional scatterings of points $x$ correspond to approximately
$1/5$ of the total of $5000$ objects (circles and squares).  Thinner slicing would 
require more objects, in order to allow more points to fall inside the slicing cylinders.
 
Table~\ref{tab:res} presents, for each of the considered features, 
the number of occurrences of successful cluster identification
(i.e.~cases in which at least two well-defined clusters were identified), the average
relevance, and the respective  product of these two values.  
  
\begin{table}
\def\arraystretch{1.5}
\caption{The number of occurrences in which clusters have been identified, 
the average relevance, and the product of these two values,
respectively to each of the 5 considered features, obtained for the 
set of 5000 circles and squares by using the
suggested one-dimensional feature selection approach. } \label{tab:res} 
\begin{center}
\begin{tabular}{|  c  |  c  | c  |  c  |}
\hline
feature  & occurrss  &  relev. $(\rho)$ &  (occurs.)(relev.)  \\
\hline
$\gamma$  & 0  &  0  &  0  \\  \hline
$perim.$  & 1  &  0.7467  &  0.7467  \\  \hline
$area$  & 2  &  0.9137  &  1.8276  \\  \hline
$rel. perim.$  & 61  &  0.3407  &  20.7849  \\  \hline
$circ.$  & 67 &  0.3553  &  23.8109  \\
\hline
\end{tabular}
\end{center}
\end{table}

All the obtained indicators suggest that the $\gamma$, $p$ and $a$ measurements were
not effective, while $r$ and $c$ yielded not only many identification occurrences,
but also a relatively high (in the order of $1/3$) height of the shorter main cluster,
leading to the highest relevance values $\rho$, which turned out to be similar.
This is in agreement with what could be expected from the type of considered shapes
and the properties of the adopted measurements, indicating the potential of the proposed
method for feature selection.

\section{Cluster Detection}

In addition to applications to feature selection, the one-dimensional slicing method reported in 
this article can also be considered for cluster detection approaches.  
The basic idea is that patches of the interstices can be identified by the one-dimensional
method and incrementally merged  so as to obtain longer, more complete separation regions.  

Many approaches can be tried, but here we consider only one approach focusing on the
identification of patches of interstices between potentially existing clusters.  The method, which
is straightforward, involves performing several one-dimensional slices and, in case clusters
are identified, to identify the respective interstice and to mark them in the original feature space.

Given an obtaine one-dimensional signal $x$ with two identified clusters,  respective interstice patches
can be estimated by identifying the limits of the two main clusters.
More specifically, the interstice can be understood as corresponding to the region between
the largest $x-$value of the left-hand side cluster and the smallest $x-$value of the
right-hand sided cluster, yielding the interval $x_{a} \leq x \leq x_b$.

\begin{figure}[h!]  
\centering{
\includegraphics[width=7cm]{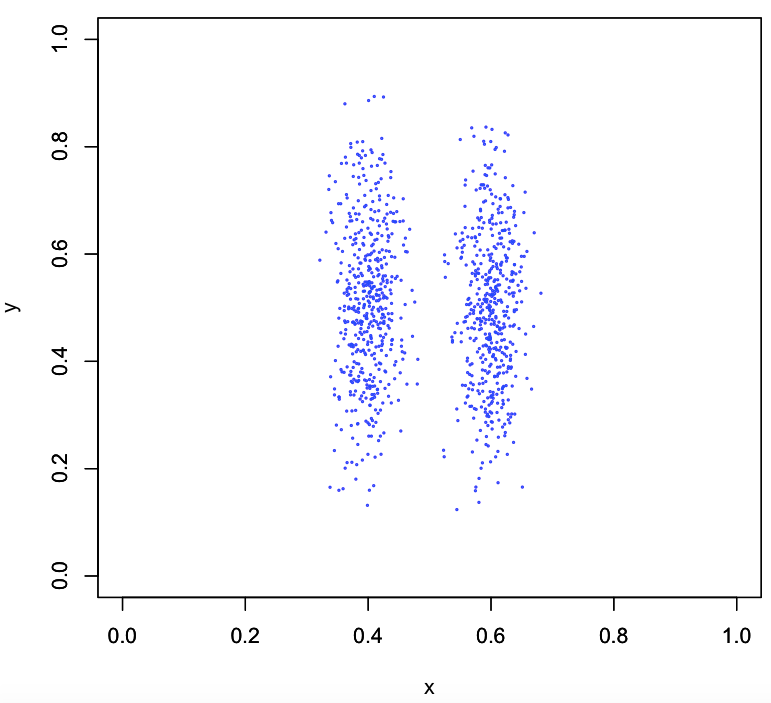}  \\
(a)   \\
\includegraphics[width=7cm]{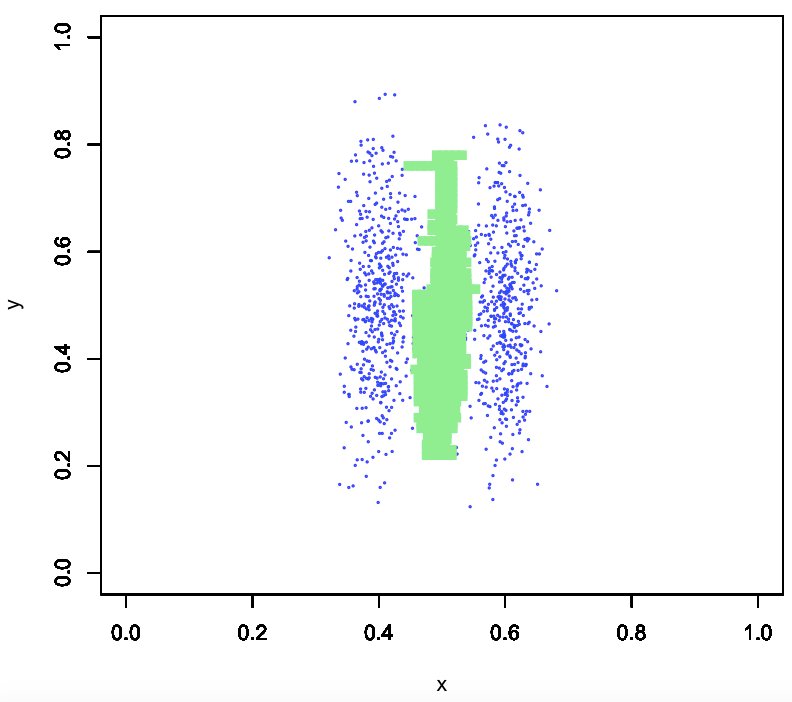}  \\
(b) \\
\caption{Two elongated clusters defined in a two-dimensional feature space (a), and
detection of the respective interstice (in light green) by using the suggested one-dimensional
clustering approach (b).}
\label{fig:2D}}
\end{figure}

In order to illustrate this simple method for interstice identification, we consider the two-dimensional
feature space shown in Figure~\ref{fig:2D}(a), containing $1000$ points organized into two
elongated clusters in a two-dimensional feature space defined by two hypothetical features
$f_1$ and $f_2$.

One-dimensional clustering was performed respectively to several slices taken along the
$x-$axis.  More specifically, we considered $y = 0, 0.01, 0.02, \ldots, 1$, $\alpha = 4$ and $r=0.1$,
while only slices containing more than 100 points were taken into account.  In each case of relevant
cluster indication, the interstice was identified and marked into the feature space.   The
result, given in Figure~\ref{fig:2D}(b), corresponds to a proper identification of the separation
region in the 2D space.

\section{Concluding Remarks}

More traditional clustering approaches, mostly oriented toward the identification of
\emph{granular} and \emph{separatrix} groups, aim at obtaining well-separated (often 
topologically) groups. These methods tend to have an inherent topologically
\emph{global} nature, in the sense of
each cluster being considered with respect to all other adjacent portions and directions of the feature space.  
Despite their potential for achieving more complete cluster identification, the global requirement 
can impose severe conceptual and computational demands on respective methods.

The present work main objective has been to extend the definition of clusters in order to include
situations where each group can be partially attached to other clusters, giving rise to the
paradoxical concept of \emph{linked cluster}. 
Such a generalized understanding of clusters immediately implies the need for methods to complement
more traditional approaches aimed at obtaining completely detached groups.  It has been
argued here that the proper patched identification of clusters and respective interstices can 
be, in principle, performed with simplicity, low computational cost, and relative efficiency by using 
the reported one-dimensional single-linkage agglomerative method.

A simple procedure has been outlined for, given a one-dimensional slice signal $x$, to look for
existence of clusters, returning as result a relevance index $0 \leq \rho \leq 1$ corresponding
to the length of the second longest dendrogram branch divided by the largest heigh value.
This algorithms was shown to perform properly for uniform and normal distribution of points
devoid of clusters, as well as for a clustering model involving two intervals with relatively
high density of points separated by a less intense density taking place along an interstice.

This simple one-dimensional method for identifying the relevance of clusters along sliced
signals was then applied to derive an algorithm for feature selection.  The basic idea was
to obtain several one-dimensional probes along each feature taken in turn, placed at varying 
configurations of the other remaining features, and to consider those features leading to higher 
relevance of clusters existence as being potentially more effective to be selected as features in
diverse clustering algorithms.   This approach was successfully illustrated with respect to
a hypothetical set of data including circles and squares with uniformly distributed radius/sides.
The suggested method was capable of identifying, among the 5 considered features, the two
alternative that were potentially more relevant for discriminating between the considered objects.

The same one-dimensional method for cluster identification in signals sliced from feature
spaces was then briefly considered as the basis for developing cluster detection algorithms.
A simple method was suggested which involves the identification of the coordinate of the
interstices along the sliced signals.   This method was successfully applied to a simple
two-dimensional clustering problem, illustrating its operation and potential.

Though illustrated with respect to relatively simple situations, the proposed methods have
potential for good performance in several types of real-world problems.  Additional 
experiments considering higher dimensional feature spaces, as well as several
types of features and number of classes and individuals, could provide a better indication about the 
characteristics of the proposed approaches.

All in all, we hoped to have described the following main contributions: (a) generalize
the concept of cluster to incorporate partial clusters; (b) to describe a simple and effective
one-dimensional method for local cluster identification in slices obtained from feature
spaces; (c) to illustrate the particular suitability of the single-linkage agglomerative
approach for this finality; (d) to derive a simple and potentially effective method for
feature selection; and (e) to outline a simple method for cluster detection.

The reported contributions pave the way to several possible future developments.  In particular,
it would be interesting to develop more sophisticate and robust cluster detection approaches,
e.g.~adapted to multiple classes, and also considering linear combinations of the original
measurements.  It would also be interesting to incorporate the described one-dimensional
clustering approach into more global pattern recognition methods possibly underlain by optimization
regarding the slicing configurations and spatial scales.   In addition, the proposed concepts
and results may also cast some light on the workings of deep learning approaches, such as
by helping to understand how  effective features can be automatically obtained.

\vspace{0.7cm}
\textbf{Acknowledgments.}

Luciano da F. Costa
thanks CNPq (grant no.~307085/2018-0) for sponsorship. This work has benefited from
FAPESP grant 15/22308-2.  
\vspace{1cm}

\bibliography{mybib}
\bibliographystyle{unsrt}

\end{document}